\documentclass[runningheads]{llncs}
\usepackage{graphicx}

\usepackage{multirow}
\usepackage{makecell}
\usepackage{tabulary}
\usepackage{booktabs}
\usepackage{bm}

\usepackage{tikz}
\usepackage{comment}
\usepackage{amsmath,amssymb} 
\usepackage{color}

\usepackage[accsupp]{axessibility}  

\begin{document}
\pagestyle{headings}
\mainmatter
\def\ECCVSubNumber{6127}  

\title{Wave-ViT: Unifying Wavelet and Transformers \\for Visual Representation Learning} 


\titlerunning{Wavelet Vision Transformer}
%
\author{Ting Yao\inst{1} \and
Yingwei Pan\inst{1} \and
Yehao Li\inst{1} \and
Chong-Wah Ngo\inst{2} \and
Tao Mei\inst{1}}
%
\authorrunning{T. Yao et al.}
%
\institute{JD Explore Academy, China \and
Singapore Management University, Singapore
\email{\{tingyao.ustc,panyw.ustc,yehaoli.sysu\}@gmail.com,cwngo@smu.edu.sg,tmei@jd.com}}
\maketitle

\begin{abstract}
Multi-scale Vision Transformer (ViT) has emerged as a powerful backbone for computer vision tasks, while the self-attention computation in Transformer scales quadratically w.r.t. the input patch number. Thus, existing solutions commonly employ down-sampling operations (\emph{e.g.}, average pooling) over keys/values to dramatically reduce the computational cost. In this work, we argue that such over-aggressive down-sampling design is not invertible and inevitably causes information dropping especially for high-frequency components in objects (\emph{e.g.}, texture details). Motivated by the wavelet theory, we construct a new Wavelet Vision Transformer (\textbf{Wave-ViT}) that formulates the invertible down-sampling with wavelet transforms and self-attention learning in a unified way. This proposal enables self-attention learning with lossless down-sampling over keys/values, facilitating the pursuing of a better efficiency-vs-accuracy trade-off. Furthermore, inverse wavelet transforms are leveraged to strengthen self-attention outputs by aggregating local contexts with enlarged receptive field. We validate the superiority of Wave-ViT through extensive experiments over multiple vision tasks (\emph{e.g.}, image recognition, object detection and instance segmentation). Its performances surpass state-of-the-art ViT backbones with comparable FLOPs. Source code is available at \url{https://github.com/YehLi/ImageNetModel}.

\end{abstract}

\section{Introduction}

Recently, leveraging Transformer architecture \cite{vaswani2017attention} for visual representation learning has achieved widespread dominance in computer vision field. Transformer architecture has brought forward milestone improvement for a series of downstream vision tasks \cite{carion2020end,chu2021twins,dosovitskiy2020image,li2022comprehending,liu2021swin,long2022stand,long2022DTF,pan2020x,wang2021pvtv2,zhang2021token,zhang2022exploring}, including both image recognition and dense prediction tasks (\emph{e.g.}, object detection and semantic segmentation). At its heart is a basic self-attention block that triggers long-range interaction among visual tokens.
The Vision Transformer (ViT) \cite{dosovitskiy2020image} is one of the early attempts that directly employs a pure Transformer over image patches, and manages to attain competitive image recognition performance against CNN counterparts. However, applying the primary ViT architecture using its outputs of single-scale and low-resolution feature map for the pixel-level dense prediction tasks (\emph{e.g.}, instance/semantic segmentation) is not trivial. Therefore, considering that visual patterns commonly occur at multiple scales in natural scenery, there has been research efforts pushing the limits of ViT backbones by aggregating contexts from multiple scales (\emph{e.g.}, ``pyramid'' strategy). For example, Pyramid Vision Transformer (PVT) \cite{wang2021pyramid,wang2021pvtv2} integrates pyramid structure into Transformer framework, yielding multi-scale feature maps for dense prediction tasks. Multiscale Vision Transformers (MViT) \cite{fan2021multiscale} learns multi-scale feature hierarchies in Transformer architecture by hierarchically expanding the channel capacity while reducing the spatial resolution.

\begin{figure*}[!tb]
\vspace{-0.2in}
\centering {\includegraphics[width=0.8\textwidth]{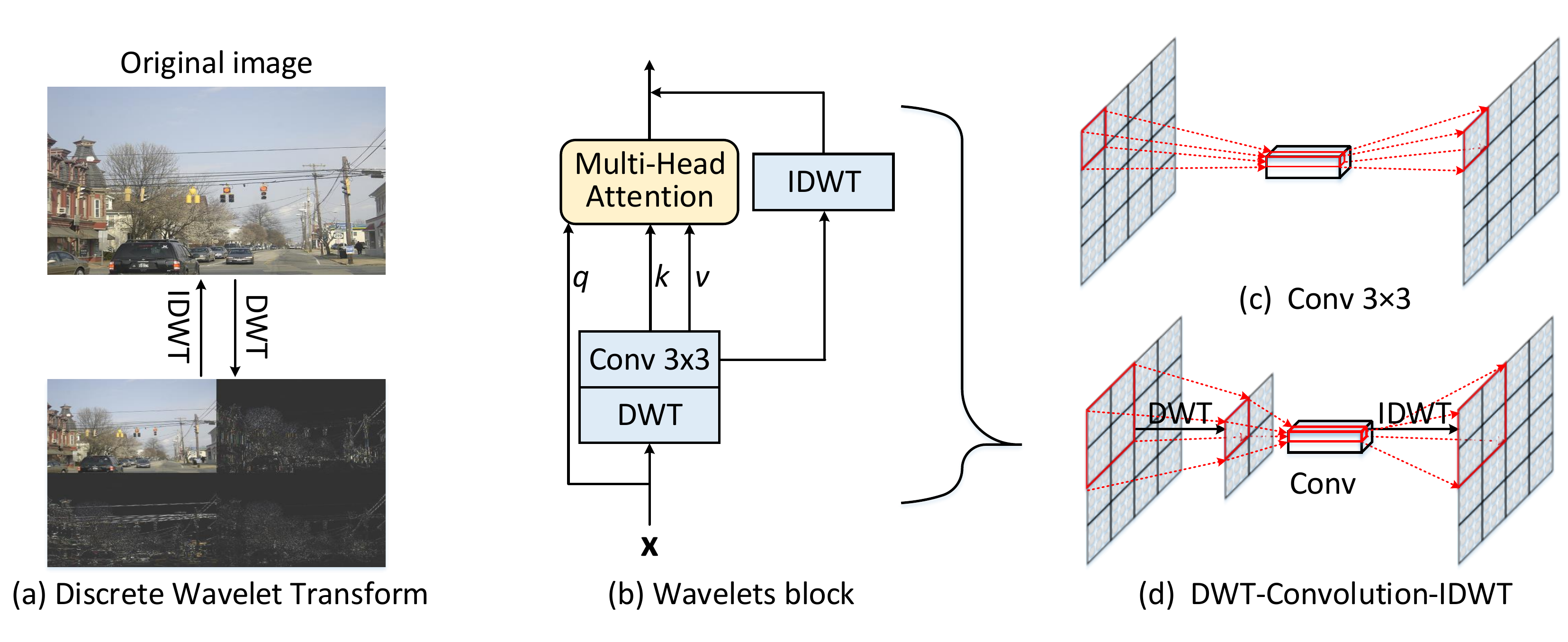}}
\vspace{-0.2in}
\caption{\small An illustration of (a) Discrete Wavelet Transform (DWT) and Inverse DWT (IDWT) over an image, (b) our Wavelets block, and the comparison between (c) a single 3$\times$3 convolution and (d) DWT-Convolution-IDWT process in our Wavelets block.}
\label{fig:fig1}
\vspace{-0.22in}
\end{figure*}

One primary challenge of applying self-attention over multi-scale feature maps is the quadratical computational cost that scales w.r.t the number of input patches (\emph{i.e.}, spatial resolution). Thus, typical multi-scale ViT approaches usually perform down-sampling operations (\emph{e.g.}, average pooling in \cite{wang2021pvtv2} or pooling kernels in \cite{fan2021multiscale}) over keys/values to reduce computational cost. Nevertheless, these pooling based operations inevitably result in information dropping (\emph{e.g.}, the high-frequency components of object texture details), and thus adversely affect the performances especially for dense prediction tasks. Furthermore, the recent studies (\emph{e.g.}, \cite{zhang2019making}) also have shown that applying pooling operations in CNNs would hurt the shift-equivariance of deep networks.

In this paper, we propose \textbf{Wavelets block} to perform invertible down-sampling through wavelet transforms, aiming to preserve the original image details for self-attention learning while reducing computational cost. Wavelet transform is a fundamental time-frequency analysis method that decomposes input signals into different frequency subbands to address the aliasing problem. In particular, Discrete Wavelet Transform (DWT) \cite{mallat1989theory} enables invertible down-sampling by transforming 2D data into four discrete wavelet subbands (Figure \ref{fig:fig1} (a)): low-frequency component ($I_{LL}$) and high-frequency components ($I_{LH}$, $I_{HL}$, $I_{HH}$). Here the low-frequency component reflects the basic object structure at coarse-grained level, while the high-frequency components retain the object texture details at fine-grained level. In this way, various levels of image details are preserved in different subbands of lower resolution without information dropping. Furthermore, inverse DWT (IDWT) can be applied to reconstruct the original image. The information preserving transformation motivates the design of an efficient Transformer block with lossless and invertible down-sampling for self-attention learning over multi-scale feature maps.

Technically, as shown in Figure \ref{fig:fig1} (b), Wavelets block first employs DWT to transform each input key/value to four subbands of lower resolution. After stacking the four subbands into a down-sampled feature map, a 3$\times$3 convolution is performed to further impose spatial locality over the frequency subbands. This leads to locally contextualized down-sampled keys/values. The multi-head self-attention learning is conducted on the down-sampled keys/values and input query. Meanwhile, IDWT can be applied over the down-sampled keys/values to reconstruct high-resolution feature map that preserves image details. Compared to the single 3$\times$3 convolution (Figure \ref{fig:fig1} (c)), the process of DWT-Convolution-IDWT (Figure \ref{fig:fig1} (d)) enables a stronger local contextualization via enlarged receptive field, with negligible increase in computation and memory. Finally, we combine the attended feature map via self-attention learning and the reconstructed feature map with local contextualization as the outputs of Wavelets~block.

By operating Wavelets block over multi-scale features in the multi-stage Transformer framework, we present a new Wavelet Vision Transformer (\textbf{Wave-ViT}) for visual representation learning.
The proposed Wave-ViT has been properly analyzed and verified through extensive experiments over different vision tasks, which demonstrate its superiority against state-of-the-art ViT backbones. More remarkably, under a comparable number of parameters, Wave-ViT achieves 85.5\% top-1 accuracy on ImageNet for image recognition, which absolutely improves PVT (83.8\%) with 1.7\%. For object detection and instance segmentation on COCO, Wave-ViT absolutely surpasses PVT with 1.3\% and 0.5\% mAP, with 25.9\% less parameters.

\section{Related Work}

\subsection{Visual Representation Learning}

Early studies in the last decade predominantly focused on exploring Convolutional Neural Networks (CNN) for visual representation learning, leading to a series of CNN backbones, \emph{e.g.}, \cite{he2016deep,krizhevsky2012imagenet,lecun1998gradient,simonyan2014very,szegedy2015going}. Most of them stack low-to-high convolutions by going deeper, targeting for producing low-resolution and high-level representations tailored for image recognition. However, dense prediction tasks like instance/semantic segmentation require high-resolution and even pixel-level representations. To tackle this, several multi-scale CNN backbones are established. For example, Res2Net \cite{gao2019res2net} presents a multi-scale building block that contains hierarchical residual-like connections. HRNet \cite{wang2020deep} connects high-to-low resolution convolution streams in parallel and meanwhile exchanges the information across resolutions repeatedly, thereby maintaining high-resolution features throughout the process.

Recently, due to the powerful long-range interaction modeling in Transformer \cite{vaswani2017attention}, Transformer has advanced natural language understanding. Inspired by this, numerous Transformer-based architectures for vision understanding have started. A few attempts augment convolutional operators with the global self-attention \cite{bello2019attention} or local self-attention \cite{hu2019local,ramachandran2019stand,srinivas2021bottleneck,zhao2020exploring}, yielding a hybrid backbone of CNN and Transformer. On a parallel note, Vision Transformer (ViT) \cite{dosovitskiy2020image} first employs a pure Transformer over the sequence of image patches for image recognition. DETR \cite{carion2020end} also leverages a pure Transformer to construct an end-to-end detector for object detection.
Different from ViT that solely divides input image into patches, TNT \cite{han2021transformer} first decomposes the inputs into several patches as ``visual sentences'', and then divides them into sub-patches as ``visual words''. A sub-transformer is additionally integrated into Transformer to perform self-attention over smaller ``visual words''.
Subsequently, to facilitate dense prediction tasks, multi-scale paradigm is introduced into Transformer structure, leading to multi-scale Vision Transformer backbones \cite{fan2021multiscale,liu2021swin,wang2021pvtv2,wang2021pyramid}. In particular, Swin Transformer \cite{liu2021swin} upgrades ViT by constructing hierarchical feature maps via merging image patches in deeper layers. Pyramid Vision Transformer (PVT) \cite{wang2021pyramid} designs a pyramid structure Transformer that produces multi-scale feature maps in a four-stage architecture. PVTv2 \cite{wang2021pvtv2} further improves PVT by using average pooling to reduce spatial dimension of keys/values, rather than convolutions in PVT. Multiscale Vision Transformers (MViT) \cite{fan2021multiscale} integrates Transformer framework with multi-scale feature hierarchies, and pooling kernels is employed over query/keys/values for spatial reduction.

Our Wave-ViT is also a type of multi-scale ViT. Existing multi-scale ViTs (\emph{e.g.}, \cite{fan2021multiscale,wang2021pvtv2,wang2021pyramid}) commonly adopt irreversible down-sampling operations like average pooling or pooling kernels for spatial reduction. In contrast, Wave-ViT capitalizes on wavelet transforms to reduce spatial dimension of keys/values via invertible down-sampling for self-attention learning over multi-scale features, leading to a better trade-off between computation cost and~performance.

\subsection{Wavelet Transform in Computer Vision}

Wavelet Transform is effective for time-frequency analysis. Considering that Wavelet Transform is invertible and capable of preserving all information, Wavelet Transform has been exploited in CNN architectures for performance boosting in various vision tasks. For example, in \cite{bae2017beyond}, Bae \emph{et al.} validate that learning CNN representations over wavelet subbands can benefit the task of image restoration. DWSR \cite{guo2017deep} takes low-resolution wavelet subbands as inputs to recover the missing details for image super-resolution task. Multi-level wavelet transform \cite{liu2018multi} is utilized to enlarge receptive field without information dropping for image restoration.
Williams \emph{et al.} \cite{williams2018wavelet} utilize Wavelet Transform to decompose input features into a second level decomposition, and discard first-level subbands to reduce feature dimensions for image recognition. Haar wavelet CNNs is integrated with multi-resolution analysis in \cite{fujieda2018wavelet} for texture classification and image annotation. In \cite{oyallon2017scaling}, ResNet is remoulded by combining the first layer with a wavelet scattering network, which achieves comparable performances on image recognition with less parameters.

Although wavelet transform has been exploited as down-sampling/up-sampling operations in CNNs, it is never explored for Transformer architecture.
In this work, our Wave-ViT goes beyond existing CNNs that operate wavelet transform over feature maps across different stages, and leverages wavelet transform to down-sample keys/values within Transformer block, making the impact more thorough for feature learning.


\section{Our Approach: Wavelet Vision Transformer}

This section starts by briefly reviewing the most typical multi-head self-attention block in ViTs, particularly on how the self-attention block is scaled down for reducing computational cost in the existing multi-scale ViTs. After that, a novel principled Transformer building block, named Wavelets block, is designed to integrate self-attention learning with wavelet transforms in a unified fashion. Such design upgrades typical self-attention block by exploiting wavelet transforms to perform invertible down-sampling, which elegantly reduces spatial dimension of keys/values without information dropping. Furthermore, this block applies inverse wavelet transforms over down-sampled keys/values to enhance outputs with enlarged receptive field. Finally, after applying Wavelets block over multi-scale features in the multi-stage Transformer architecture, we elaborate a new multi-scale ViT backbone, \emph{i.e.}, Wavelet Vision Transformer.

\begin{figure*}[!tb]
\vspace{-0.1in}
\centering {\includegraphics[width=0.9\textwidth]{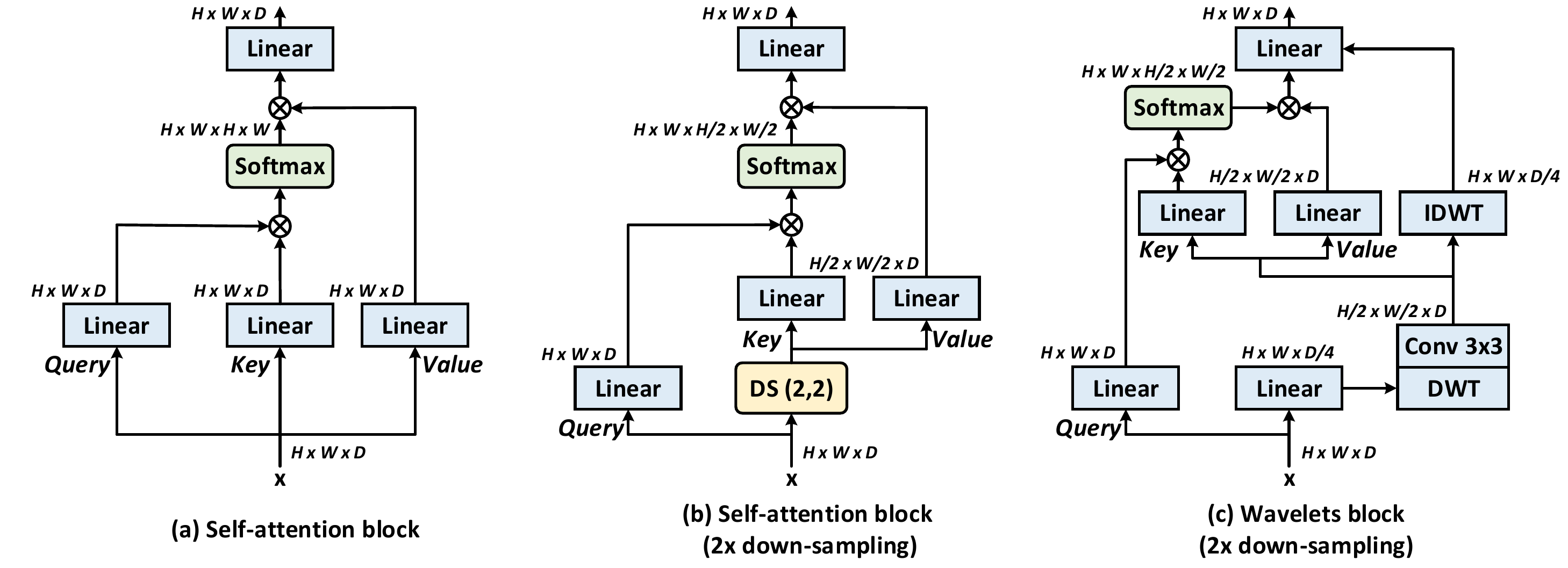}}
\vspace{-0.2in}
\caption{\small The detailed architectures of (a) basic self-attention block in ViT Backbones, (b) self-attention block with down-sampling operation (\emph{i.e.}, DS$(2,2)$) that reduces the spatial scale of both height and width by half, and (c) our Wavelets block that capitalizes on wavelet transforms to enable lossless down-sampling.}
\label{fig:framework}
\vspace{-0.2in}
\end{figure*}

\subsection{Preliminaries}

\subsubsection{Multi-head Self-attention in ViT Backbones.}
Mainstream Transformer architectures, especially Vision Transformer backbones \cite{dosovitskiy2020image}, often rely on the typical multi-head self-attention that captures long-range dependencies among inputs in a scalable fashion. Here we present a general formulation of multi-head self-attention as illustrated in Figure \ref{fig:framework} (a). Technically, let $X \in {{\mathbb{R}}^{H \times W \times D}}$ be the input 2D feature map, where $H$/$W$/$D$ denote the height/width/channel number, respectively. Here $X$ can be reshaped as a patch sequence, consisting of $n = H \times W$ image patches and the dimension of each patch is $D$. We linearly transform the input patch sequence $X$ into three groups in parallel: queries $Q \in {{\mathbb{R}}^{n \times D}}$, keys $K \in {{\mathbb{R}}^{n \times D}}$, and values $V \in {{\mathbb{R}}^{n \times D}}$. After that, the multi-head self-attention ({\bf{MultiHead}}) module \cite{vaswani2017attention} decomposes each query/key/value into $N_h$ parts along channel dimension, leading to queries ${Q_j} \in {{\mathbb{R}}^{n \times {D_h}}}$, keys ${K_j} \in {{\mathbb{R}}^{n \times {D_h}}}$, and values ${V_j} \in {{\mathbb{R}}^{n \times {D_h}}}$ for the $j$-th head. Note that $N_h$ is head number and $D_h$ denotes the dimension of each head. Then, we perform self-attention learning ({\bf{Attention}}) over queries, keys and values for each head, and the outputs of each head are concatenated, followed by a linear transformation to compose the final outputs:
\begin{equation}\small
\label{eq:sa1}
\begin{aligned}
&{{\bf{MultiHead}}(Q,K,V) = {\bf{Concat}}(head_0, head_1,...,head_{N_h})W^O},\\
&{head_j = {\bf{Attention}}(Q_j,K_j,V_j)},\\
&{{\bf{Attention}}(Q_j,K_j,V_j) = {\bf{Softmax}}(\frac{{{Q_j}{{K_j}^T}}}{{\sqrt {D_h} }}){V_j}},
\end{aligned}
\end{equation}
where ${\bf{Concat}}(\cdot)$ is the concatenation operation and $W^O$ is the transformation matrix.
According to the general formulation in Eq.(\ref{eq:sa1}), the computational cost of multi-head self-attention for the input feature map $X \in {{\mathbb{R}}^{H \times W \times D}}$ is $\mathcal{O}(H^2W^2D)$, which scales quadratically w.r.t. the input patch number. In this way, such design inevitably leads to a sharp rise in computational cost especially for high-resolution inputs.

\subsubsection{Self-attention with Down-sampling in Multi-scale ViT Backbones.}
To alleviate the heavy self-attention computation overhead for high-resolution inputs, the existing multi-scale ViT backbones commonly adopt the down-sampling operations (\emph{e.g.}, average pooling in \cite{wang2021pvtv2} or pooling kernels in \cite{fan2021multiscale}) over keys/values for spatial reduction. Taking the self-attention block with 2$\times$ down-sampling in Figure \ref{fig:framework} (b) as an example, the input 2D feature map $X$ is first down-sampled by a factor $r$ ($r=2$ in this case). Here the down-sampling operator is denoted as DS$(2,2)$, that reduces the spatial scale of both height and width by half. Next, the down-sampled feature map is linearly transformed into keys $K^d \in {{\mathbb{R}}^{{\frac{n}{r^2}} \times D}}$ and values $V^d \in {{\mathbb{R}}^{{\frac{n}{r^2}} \times D}}$ to trigger multi-head self-attention learning. As such, the overall computational cost of multi-head self-attention is dramatically reduced by a factor of $r^2$ (\emph{i.e.}, $\mathcal{O}(\frac{H^2W^2D}{r^2})$).

\subsection{Wavelets Block}
Although the aforementioned multi-scale ViT backbones reduce self-attention computation via down-sampling, the commonly adopted down-sampling operations like average pooling are irreversible, and inevitably result in information dropping. To mitigate this issue, we design a principled self-attention block, named \textbf{Wavelets block}, that novelly capitalizes on wavelet transforms to enable invertible down-sampling for self-attention learning. Such invertible down-sampling is seamlessly incorporated into the typical self-attention block, pursuing efficient multi-head self-attention learning with lossless down-sampling. Figure \ref{fig:framework} (c) details the architecture of our Wavelets block.

Formally, given the input 2D feature map $X \in {{\mathbb{R}}^{H \times W \times D}}$, we first linearly transform it into $\widetilde X = XW_d$ with reduced channel dimension via embedding matrix $W_d \in {{\mathbb{R}}^{D \times {\frac{D}{4}}}} $. Next, we employ Discrete Wavelet Transform (DWT) to down-sample the input $\widetilde X \in {{\mathbb{R}}^{H \times W \times {\frac{D}{4}}}}$ by decomposing it into four wavelet subbands. Note that here we choose the classical Haar wavelet for DWT as in \cite{liu2020wavelet} for simplicity. Concretely, DWT applies the low-pass filter $f_L = (1/\sqrt 2, 1/\sqrt 2)$ and high-pass filter $f_H = (1/\sqrt 2, -1/\sqrt 2)$ along the rows to encode $\widetilde X$ into two subbands $X_L$ and $X_H$. Next, the same low-pass filter $f_L$ and high-pass filter $f_H$ are employed along the columns of the learnt subbands $X_L$ and $X_H$, leading to all the four wavelet subbands: $X_{LL}\in {{\mathbb{R}}^{{\frac{H}{2}} \times {\frac{W}{2}} \times {\frac{D}{4}}}}$, $X_{LH}\in {{\mathbb{R}}^{{\frac{H}{2}} \times {\frac{W}{2}} \times {\frac{D}{4}}}}$, $X_{HL}\in {{\mathbb{R}}^{{\frac{H}{2}} \times {\frac{W}{2}} \times {\frac{D}{4}}}}$, and $X_{HH}\in {{\mathbb{R}}^{{\frac{H}{2}} \times {\frac{W}{2}} \times {\frac{D}{4}}}}$. $X_{LL}$ refers to the low-frequency component that reflects the basic object structure at coarse-grained level. $X_{LH}$, $X_{HL}$, and $X_{HH}$ represent the high-frequency components that retain the object texture details at fine-grained level. In this way, each wavelet subband can be regarded as the down-sampled version of $\widetilde X$, and all of them cover every detail of inputs without any information dropping.

We concatenate the four wavelet subbands along the channel dimension to form $\hat X=[X_{LL},X_{LH},X_{HL},X_{HH}]\in {{\mathbb{R}}^{{\frac{H}{2}} \times {\frac{W}{2}} \times D}}$. A 3$\times$3 convolution is further applied to impose spatial locality over $\hat X$, yielding the locally contextualized down-sampled feature map $X^c$. Next, this down-sampled feature map $X^c$ is linearly transformed into down-sampled keys $K^{w} \in {^{m \times D}}$ and values $V^{w} \in {^{m \times D}}$, where $ m = {\frac{H}{2}} \times {\frac{W}{2}} $ is the number of patches. Similarly, the wavelet-based multi-head self-attention learning ${\bf{Attention^w}}$ is thus performed over the queries and the corresponding down-sampled keys/values for each head:
\begin{equation}\small
\label{eq:wavelets_sa1}
\begin{aligned}
head_j={{\bf{Attention^w}}(Q_j,K^w_j,V^w_j) = {\bf{Softmax}}(\frac{{{Q_j}{{K^w_j}^T}}}{{\sqrt {D_h} }}){V^w_j}},
\end{aligned}
\end{equation}
where $K^w_j$/$V^w_j$ denotes the down-sampled keys/values for the $j$-th head, respectively. Here the aggregated output of self-attention learning for each head ($head_j$) can be interpreted as the long-range contextualized information of~inputs.

As a beneficial by-product, we additionally apply inverse DWT (IDWT) over the locally contextualized down-sampled feature $X^c$. According to the wavelet theory, the reconstructed feature map $X^r$ is able to retain every detail of primary input $\widetilde X$. It is worthy to note that compared to a single 3$\times$3 convolution, such process of DWT-Convolution-IDWT in Wavelets block triggers a stronger local contextualization with enlarged receptive field, with negligible increase in computational cost/memory.

Finally, we concatenate all the long-range contextualized information of each head plus the reconstructed locally contextualized information $X^r$, followed by a linear transformation to compose the outputs of our Wavelets block:
\begin{equation}\small
\label{eq:block}
\begin{aligned}
&{{\bf{WaveletsBlock}}(X) = {\bf{MultiHead^w}}(X{W^q},{X^c}{W^k},{X^c}{W^v},X^r)},\\
&{{\bf{MultiHead^w}}(Q,K,V,X^r) = {\bf{Concat}}(head_0, head_1,...,head_{N_h}, X^r) \widetilde{W}^O},\\
\end{aligned}
\end{equation}
where $\widetilde{W}^O$ is the transformation matrix.

\subsection{Wavelet Vision Transformer}

Recall that our Wavelets block is a principled unified self-attention block, it is feasible to construct multi-scale ViT backbones with Wavelets blocks. Following the basic configuration of existing multi-scale ViTs \cite{liu2021swin,wang2021pyramid}, we present three variants of our Wavelet Vision Transformer (Wave-ViT) with different model sizes, \emph{i.e.}, Wave-ViT-S (small size), Wave-ViT-B (base size), and Wave-ViT-L (large size). Note that Wave-ViT-S/B/L shares similar model size and computational complexity with Swin-T/S/B \cite{liu2021swin}. Specifically, given the input image (size: 224 $\times$ 224), the entire architecture of Wave-ViT consists of four stages, and each stage is comprised of a patch embedding layer, and a stack of Wavelets blocks followed by feed-forward layers. We follow the design principle of ResNet \cite{he2016deep} by progressively increasing the channel dimensions of all the four stages and meanwhile shrinking the spatial resolutions. Table \ref{table:architecture} details the architectures of all the three variants of Wave-ViT, where $E_i$, $Head_i$, and $C_i$ is the expansion ratio of feed-forward layer, head number, and the channel dimension in stage~$i$.

\begin{table*}[!tb]\scriptsize
\centering
\caption{Detailed architecture specifications for three variants of our Wave-ViT with different model sizes, \emph{i.e.}, Wave-ViT-S (small size), Wave-ViT-B (base size), and Wave-ViT-L (large size). $E_i$, $Head_i$, and $C_i$ represents the expansion ratio of feed-forward layer, the head number, and the channel dimension in each stage $i$, respectively.}
\vspace{-0.1in}
\begin{tabular}{c|c|c|c|c}
\Xhline{2\arrayrulewidth}
        & Output Size & Wave-ViT-S & Wave-ViT-B & Wave-ViT-L \\ \hline
Stage 1 & $\frac{H}{4} \times \frac{W}{4}$
        & $\left[ \begin{array}{c}  E_1=8 \\ Head_1=2 \\ C_1=64  \end{array} \right] \!\times\! 3$
        & $\left[ \begin{array}{c}  E_1=8 \\ Head_1=2 \\ C_1=64  \end{array} \right] \!\times\! 3$
        & $\left[ \begin{array}{c}  E_1=8 \\ Head_1=3 \\ C_1=96  \end{array} \right] \!\times\! 3$
        \\ \hline
Stage 2 & $\frac{H}{8} \times \frac{W}{8}$
        & $\left[ \begin{array}{c} E_2=8 \\ Head_2=4 \\  C_2=128 \end{array} \right] \!\times\! 4$
        & $\left[ \begin{array}{c} E_2=8 \\ Head_2=4 \\  C_2=128 \end{array} \right] \!\times\! 4$
        & $\left[ \begin{array}{c} E_2=8 \\ Head_2=6 \\  C_2=192 \end{array} \right] \!\times\! 6$
        \\ \hline
Stage 3 & $\frac{H}{16} \times \frac{W}{16}$
        & $\left[ \begin{array}{c}  E_3=4 \\ Head_3=10 \\ C_3=320 \end{array} \right] \!\times\! 6$
        & $\left[ \begin{array}{c}  E_3=4 \\ Head_3=10 \\ C_3=320 \end{array} \right] \!\times\! 12$
        & $\left[ \begin{array}{c}  E_3=4 \\ Head_3=12 \\ C_3=384 \end{array} \right] \!\times\! 18$
        \\ \hline
Stage 4 & $\frac{H}{32} \times \frac{W}{32}$
        & $\left[ \begin{array}{c} E_4=4 \\ Head_4=14 \\ C_4=448 \end{array} \right] \!\times\! 3$
        & $\left[ \begin{array}{c} E_4=4 \\ Head_4=16 \\ C_4=512 \end{array} \right] \!\times\! 3$
        & $\left[ \begin{array}{c} E_4=4 \\ Head_4=16 \\ C_4=512 \end{array} \right] \!\times\! 3$
        \\ \Xhline{2\arrayrulewidth}
\end{tabular}
\label{table:architecture}
\vspace{-0.2in}
\end{table*}

\section{Experiments}
We evaluate the effectiveness of our proposed multi-scale ViT backbone, denoted as Wave-ViT, through various empirical evidence on a series of mainstream CV tasks, including image recognition, object detection, instance segmentation, and semantic segmentation. Concretely, we consider the following evaluations to compare the quality of learnt feature representations obtained from various vision backbones: (a) Training from scratch for image recognition task on ImageNet1K \cite{deng2009imagenet}; (b) Fine-tuning the backbones (pre-trained on ImageNet1K) for downstream tasks, \emph{i.e.}, object detection and instance segmentation on COCO \cite{lin2014microsoft}, and semantic segmentation on ADE20K \cite{zhou2019semantic}; (c) Ablation studies that support each design in our Wavelets block; (d) Visualization of learnt visual representation by Wave-ViT.

\begin{table*}[!tb]\scriptsize
\centering
\vspace{-0.1in}
\caption{The performances of various vision backbones on ImageNet1K dataset for image recognition task. $\star$ indicates that the vision backbone is additionally trained with Token Labeling objective with MixToken \cite{jiang2021all} and convolutional stem (conv-stem) \cite{wang2021scaled} for patch encoding. We group the vision backbones into three categories, and all backbones within each category shares similar GFLOPs: Small (GFLOPs$<$6), Base (6$\leq$GFLOPs$<$10), Large (10$\leq$GFLOPs$<$22).}
\vspace{-0.1in}
\setlength{\tabcolsep}{0.1pt}
\begin{tabular}{c|c|c|cc|c|cc|cc}
\Xhline{2\arrayrulewidth}
Method          & Params & GFLOPs & Top-1 & Top-5 & Method          & Params & GFLOPs & Top-1 & Top-5 \\ \hline
\multicolumn{5}{c|} {Small} & \multicolumn{5}{c} {Large} \\ \hline
ResNet-50  \cite{he2016deep}                 & 25.5M & 4.1 & 78.3 & 94.3 & ResNet-152 \cite{he2016deep}                 & 60.2M & 11.6 & 81.3 & 95.5  \\
BoTNet-S1-50 \cite{srinivas2021bottleneck}   & 20.8M & 4.3 & 80.4 & 95.0 & ResNeXt101 \cite{xie2017aggregated}    & 83.5M & 15.6 & 81.5 & -     \\
Swin-T  \cite{liu2021swin}                   & 29.0M & 4.5 & 81.2 & 95.5 & DeiT-B \cite{touvron2021training}            & 86.6M & 17.6 & 81.8 & 95.6  \\
ConViT-S \cite{d2021convit}                  & 27.8M & 5.4 & 81.3 & 95.7 & SE-ResNet-152 \cite{hu2018squeeze}           & 66.8M & 11.6 & 82.2 & 95.9  \\
T2T-ViT-14 \cite{yuan2021tokens}             & 21.5M & 4.8 & 81.5 & 95.7 & ResNeSt-101 \cite{zhang2020resnest}          & 48.3M & 10.2 & 82.3 & -     \\
RegionViT-Ti+ \cite{chen2021regionvit}       & 14.3M & 2.7 & 81.5 & -    & ConViT-B \cite{d2021convit}                  & 86.5M & 16.8 & 82.4 & 95.9  \\
SE-CoTNetD-50  \cite{li2022contextual} & 23.1M & 4.1 & 81.6 & 95.8    & T2T-ViTt-24 \cite{yuan2021tokens}            & 64.1M & 15.0 & 82.6 & 95.9  \\
Twins-SVT-S \cite{chu2021twins}              & 24.1M & 2.9 & 81.7 & 95.6 & TNT-B  \cite{han2021transformer}             & 65.6M & 14.1 & 82.9 & 96.3  \\
CoaT-Lite Small \cite{xu2021co}              & 20.0M & 4.0 & 81.9 & 95.5 & DeepViT-L  \cite{zhou2021deepvit}            & 58.9M & 12.8 & 83.1 & -     \\
PVTv2-B2 \cite{wang2021pvtv2}                & 25.4M & 4.0 & 82.0 & 96.0 & RegionViT-B \cite{chen2021regionvit}         & 72.7M & 13.0 & 83.2 & 96.1  \\
Wave-ViT-S                              & 19.8M & 4.3 & 82.7 & 96.2 & CaiT-S36 \cite{touvron2021going}             & 68.4M  & 13.9 & 83.3 & -     \\
Wave-ViT-S$^\star$                   & 22.7M & 4.7 & \textbf{83.9} & \textbf{96.6} & CrossViT-15-384 \cite{chen2021crossvit}      & 28.5M & 21.4 & 83.5 & -     \\ \cline{1-5}
\multicolumn{5}{c|} {Base}                                        & BoTNet-S1-128 \cite{srinivas2021bottleneck}  & 75.1M & 19.3 & 83.5 & 96.5  \\  \cline{1-5}
ResNet-101 \cite{he2016deep}                 & 44.6M & 7.9 & 80.0 & 95.0 & Swin-B  \cite{liu2021swin}                   & 88.0M & 15.4 & 83.5 & 96.5  \\
BoTNet-S1-59 \cite{srinivas2021bottleneck}   & 33.5M & 7.3 & 81.7 & 95.8 & PVTv2-B4 \cite{wang2021pvtv2}                & 62.6M & 10.1 & 83.6 & 96.7  \\
T2T-ViT-19 \cite{yuan2021tokens}             & 39.2M & 8.5 & 81.9 & 95.7 & Twins-SVT-L \cite{chu2021twins}              & 99.3M & 15.1 & 83.7 & 96.5  \\
CvT-21 \cite{wu2021cvt}                      & 32.0M & 7.1 & 82.5 & -    & RegionViT-B+ \cite{chen2021regionvit}        & 73.8M & 13.6 & 83.8 & -     \\
Swin-S  \cite{liu2021swin}                   & 50.0M & 8.7 & 83.2 & 96.2 & Focal-Base \cite{yang2021focal}              & 89.8M & 16.0 & 83.8 & 96.5  \\
Twins-SVT-B \cite{chu2021twins}              & 56.1M & 8.6 & 83.2 & 96.3 & PVTv2-B5 \cite{wang2021pvtv2}                & 82.0M & 11.8 & 83.8 & 96.6  \\
SE-CoTNetD-101 \cite{li2022contextual}     & 40.9M & 8.5   & 83.2 & 96.5 & SE-CoTNetD-152 \cite{li2022contextual} & 55.8M & 17.0 & 84.0 & 97.0 \\
PVTv2-B3 \cite{wang2021pvtv2}                & 45.2M & 6.9 & 83.2 & 96.5 & LV-ViT-M$^\star$ \cite{jiang2021all}         & 55.8M & 16.0 & 84.1 & 96.7  \\
RegionViT-M+ \cite{chen2021regionvit}        & 42.0M & 7.9 & 83.4 & -    & VOLO-D2$^\star$ \cite{yuan2021volo}          & 58.7M & 14.1 & 85.2 & -     \\
VOLO-D1$^\star$  \cite{yuan2021volo}         & 26.6M & 6.8 & 84.2 & - & VOLO-D3$^\star$ \cite{yuan2021volo}          & 86.3M & 20.6 & 85.4 & -     \\
Wave-ViT-B$^\star$                                   & 33.5M & 7.2 & \textbf{84.8} & \textbf{97.1} & Wave-ViT-L$^\star$   & 57.5M & 14.8 & \textbf{85.5} & \textbf{97.3}  \\
\Xhline{2\arrayrulewidth}
\end{tabular}
\label{table:imagenet}
\vspace{-0.3in}
\end{table*}

\subsection{Image Recognition on ImageNet1K}
\subsubsection{Dataset and Optimization Setups.} In the task of image recognition, we adopt the ImageNet1K benchmark, which comprises 1.28 million training images and 50K validation images from 1,000 classes. All vision backbones are trained from scratch on the training set, and both top-1 and top-5 accuracies metrics are used to evaluate the trained backbones on the validation set. During training, we follow the setups in \cite{yuan2021volo} by applying RandAug \cite{cubuk2020randaugment}, CutOut \cite{zhong2020random}, and Token Labeling objective with MixToken \cite{jiang2021all} for data augmentation. We adopt the AdamW optimizer \cite{loshchilov2017decoupled} with a momentum of 0.9. In particular, the optimization process includes 10 epochs of linear warm-up and 300 epochs with cosine decay learning rate scheduler \cite{loshchilov2016sgdr}. The batch size is set as 1,024, which is distributed on 8 V100 GPUs. We fix the learning rate and weight decay as 0.001 and 0.05.

\subsubsection{Performance Comparison.}

Table \ref{table:imagenet} summarizes the performance comparisons between the state-of-the-art vision backbones and our Wave-ViT variants. Note that the most competitive ViT backbones VOLO variants (\emph{i.e.}, VOLO-D1$^\star$, VOLO-D2$^\star$, and VOLO-D3$^\star$) are trained with additional Token Labeling objective with MixToken \cite{jiang2021all} and convolutional stem (conv-stem) \cite{wang2021scaled} for better patch encoding. We also adopt the same upgraded strategies to train our Wave-ViT, yielding the variants in each size (\emph{i.e.}, Wave-ViT-S$^\star$, Wave-ViT-B$^\star$, Wave-ViT-L$^\star$). Moreover, for fair comparison with other vision backbones without these strategies, we also implement a degraded version of Wave-ViT in Small size without Token Labeling objective and conv-stem (\emph{i.e.}, Wave-ViT-S). As shown in this table, under the similar GFLOPs for each group, our Wave-ViT variants consistently achieve better performances against the existing vision backbones, including both CNN backbones (\emph{e.g.}, ResNet and SE-ResNet), single-scale ViTs (\emph{e.g.}, TNT, CaiT, and CrossViT), and multi-scale ViTs (\emph{e.g.}, Swin, Twins-SVT, PVTv2, VOLO). In particular, under the Base size, the Top-1 accuracy score of Wave-ViT-B$^\star$ can reach 84.8\%, which leads to the absolute improvement of 0.6\% against the best competitive VOLO-D1$^\star$ (Top-1 accuracy: 84.2\%). Moreover, when removing the upgraded strategies as in VOLO for training, our Wave-ViT-S still manages to outperform the best multi-scale ViT in Small size (PVTv2-B2). These results generally demonstrate the key advantage of unifying self-attention learning and invertible down-sampling with wavelet transforms to facilitate visual representation learning. Most specifically, under the same Large size, compared to ResNet-152 and SE-ResNet-152 that solely capitalize on CNN architectures, the single-scale ViTs (\emph{e.g.}, TNT-B, CaiT-S36, and CrossViT-15-384) outperform them by capturing long-range dependency via Transformer structure. However, the performances of CaiT-S36 and CrossViT-15-384 are still lower than most multi-scale ViTs (PVTv2-B5 and VOLO-D3$^\star$) that aggregates multi-scale contexts for image recognition. Furthermore, instead of using irreversible down-sampling for self-attention learning in PVTv2-B5, our Wave-ViT-L$^\star$ enables invertible down-sampling with wavelet transforms, and thus achieves better efficiency-vs-accuracy trade-off. It is worthy to note that VOLO-D3$^\star$ does not employ down-sampling operations to reduce computational cost for high-resolution inputs, but instead directly reduces the input resolution (28$\times$28) at initial stage. In contrast, Wave-ViT-L$^\star$ keeps the high-resolution inputs (56$\times$56), and exploits wavelet transforms to trigger lossless down-sampling for multi-scale self-attention learning, leading to performance boosts.

\subsection{Object Detection and Instance Segmentation on COCO}
\subsubsection{Dataset and Optimization Setups.} In this section, we examine the pre-trained Wave-ViT's behavior on COCO dataset for two downstream tasks that localize objects ranging from bounding-box level to pixel level, \emph{i.e.}, object detection and instance segmentation. Two mainstream detectors, \emph{i.e.}, RetinaNet \cite{lin2017focal} and Mask R-CNN \cite{he2017mask}, are employed for each downstream task, and we replace the CNN backbones in each detector with our Wave-ViT for evaluation. Specifically, each vision backbone is first pre-trained over ImageNet1K, and the newly added layers are initialized with Xavier \cite{glorot2010understanding}. Next, we follow the standard setups in \cite{liu2021swin} to train all models on the COCO train2017 ($\sim$118K images). Here the batch size is set as 16, and AdamW \cite{loshchilov2017decoupled} is utilized for optimization (weight decay: 0.05, initial learning rate: 0.0001). All models are finally evaluated on the COCO val2017 (5K images). For the downstream task of object detection, we report the Average Precision($AP$) at different IoU thresholds and for three different object sizes (\emph{i.e.}, small, medium, large (S/M/L)). For the downstream task of instance segmentation, both bounding box and mask Average Precision (\emph{i.e.}, $AP^b$, $AP^m$) are reported. During training, we resize each input training image by fixing the shorter side as 800 pixels and meanwhile making the longer side not exceeding 1,333 pixels. Note that for RetinaNet and Mask R-CNN, 1 $\times$ training schedule (\emph{i.e.}, 12 epochs) is adopted to train the two mainstream detectors. In addition to RetinaNet, we also include four state-of-the-arts detectors (GFL \cite{li2020generalized}, Sparse RCNN \cite{sun2021sparse}, Cascade Mask R-CNN \cite{cai2018cascade}, and ATSS \cite{zhang2020bridging}) for object detection task. Following \cite{liu2021swin,wang2021pvtv2}, we utilize 3 $\times$ schedule (\emph{i.e.}, 36 epochs) with multi-scale strategy for training, and the shorter side of each input image is randomly resized within the range of [480, 800] while the longer side is forced to be less than 1,333 pixels.

\begin{table*}[!tb]\scriptsize
\vspace{-0.1in}
\centering
\caption{The performances of various vision backbones on COCO val2017 dataset for the downstream tasks of object detection and instance segmentation. For object detection task, we employ RetinaNet as the object detector, and the Average Precision($AP$) at different IoU thresholds or three different object sizes (\emph{i.e.}, small, medium, large (S/M/L)) are reported for evaluation. For instance segmentation task, we adopt Mask R-CNN as the base model, and the bounding box and mask Average Precision (\emph{i.e.}, $AP^b$ and $AP^m$) are reported for evaluation. We group all vision backbones into two categories: Small size and Base size.}
\vspace{-0.1in}
\setlength{\tabcolsep}{0.5pt}
\begin{tabular}{c|cccccc|cccccc}
\Xhline{2\arrayrulewidth}
\multirow{2}{*}{Backbone} & \multicolumn{6}{c|}{RetinaNet 1x \cite{lin2017focal}}       & \multicolumn{6}{c}{Mask R-CNN 1x \cite{he2017mask}}           \\ \cline{2-13}
   & $AP$ & $AP_{50}$ & $AP_{75}$ & $AP_S$ & $AP_M$ & $AP_L$ & $AP^b$ & $AP^b_{50}$ & $AP^b_{75}$ & $AP^m$  & $AP^m_{50}$ & $AP^m_{75}$ \\ \hline
ResNet50 \cite{he2016deep}                   & 36.3 & 55.3 & 38.6 & 19.3 & 40.0 & 48.8 & 38.0 & 58.6  & 41.4  & 34.4 & 55.1  & 36.7  \\
Swin-T   \cite{liu2021swin}                  & 41.5 & 62.1 & 44.2 & 25.1 & 44.9 & 55.5 & 42.2 & 64.6  & 46.2  & 39.1 & 61.6  & 42.0  \\
Twins-SVT-S \cite{chu2021twins}              & 43.0 & 64.2 & 46.3 & 28.0 & 46.4 & 57.5 & 43.4 & 66.0  & 47.3  & 40.3 & 63.2  & 43.4  \\
RegionViT-S \cite{chen2021regionvit}         & 43.9 & -    & -    & -    & -    & -    & 44.2 & -     & -     & 40.8 & -     & -     \\
PVTv2-B2  \cite{wang2021pvtv2}               & 44.6 & 65.6 & 47.6 & 27.4 & 48.8 & 58.6 & 45.3 & 67.1  & 49.6  & 41.2 & 64.2  & 44.4  \\
Wave-ViT-S                                   & \textbf{45.8} & \textbf{67.0} & \textbf{49.4} & \textbf{29.2} & \textbf{50.0} & \textbf{60.8} & \textbf{46.6} & \textbf{68.7}  & \textbf{51.2}  & \textbf{42.4} & \textbf{65.5}  & \textbf{45.8}  \\ \hline \hline
ResNet101 \cite{he2016deep}                  & 38.5 & 57.8 & 41.2 & 21.4 & 42.6 & 51.1 & 40.4 & 61.1  & 44.2  & 40.4 & 61.1  & 44.2  \\
ResNeXt101-32x4d  \cite{xie2017aggregated}   & 39.9 & 59.6 & 42.7 & 22.3 & 44.2 & 52.5 & 41.9 & 62.5  & 45.9  & 37.5 & 59.4  & 40.2  \\
Swin-S \cite{liu2021swin}                    & 44.5 & 65.7 & 47.5 & 27.4 & 48.0 & 59.9 & 44.8 & 66.6  & 48.9  & 40.9 & 63.4  & 44.2  \\
Twins-SVT-B \cite{chu2021twins}              & 45.3 & 66.7 & 48.1 & 28.5 & 48.9 & 60.6 & 45.2 & 67.6  & 49.3  & 41.5 & 64.5  & 44.8  \\
RegionViT-B \cite{chen2021regionvit}         & 44.6 & -    & -    & -    & -    & -    & 45.4 & -     & -     & 41.6 & -     & -     \\
PVTv2-B3  \cite{wang2021pvtv2}               & 45.9 & 66.8 & 49.3 & 28.6 & 49.8 & 61.4 & 47.0 & 68.1  & 51.7  & 42.5 & 65.7  & 45.7  \\
Wave-ViT-B                                   & \textbf{47.2} & \textbf{68.2} & \textbf{50.9} & \textbf{29.7} & \textbf{51.4} & \textbf{62.3} & \textbf{47.6} & \textbf{69.1}  & \textbf{52.4}  & \textbf{43.0} & \textbf{66.4}  & \textbf{46.0}  \\ \Xhline{2\arrayrulewidth}
\end{tabular}
\label{table:od}
\vspace{-0.1in}
\end{table*}

\begin{table*}[!tb]\scriptsize
\centering
\caption{The performances of various vision backbones on COCO val2017 dataset for the downstream task of object detection. Four kinds of object detectors, \emph{i.e.}, GFL \cite{li2020generalized}, Sparse RCNN \cite{sun2021sparse}, Cascade Mask R-CNN \cite{cai2018cascade}, and ATSS \cite{zhang2020bridging} in mmdetection \cite{chen2019mmdetection}, are adopted for evaluation. We report the bounding box Average Precision ($AP^b$) in different IoU thresholds.}
\vspace{-0.1in}
\begin{tabular}{c|c|ccc|c|c|ccc}
\Xhline{2\arrayrulewidth}
Backbone   & Method & $AP^b$ & $AP^b_{50}$ & $AP^b_{75}$ & Backbone   & Method   & $AP^b$ & $AP^b_{50}$ & $AP^b_{75}$ \\ \hline
ResNet50 \cite{he2016deep}  & \multirow{4}{*}{GFL \cite{li2020generalized}}
   & 44.5 & 63.0  & 48.3  & ResNet50 \cite{he2016deep} & \multirow{4}{*}{\begin{tabular}[c]{@{}c@{}}Sparse\\ R-CNN \end{tabular} \cite{sun2021sparse}} & 44.5 & 63.4  & 48.2  \\
Swin-T   \cite{liu2021swin}     &    & 47.6 & 66.8  & 51.7  & Swin-T  \cite{liu2021swin}     &         & 47.9 & 67.3  & 52.3  \\
PVTv2-B2  \cite{wang2021pvtv2}  &    & 50.2 & 69.4  & 54.7  & PVTv2-B2 \cite{wang2021pvtv2}  &         & 50.1 & 69.5  & 54.9  \\
Wave-ViT-S                 &  & \textbf{50.9} & \textbf{70.2} & \textbf{55.4} & Wave-ViT-S &  & \textbf{50.7} & \textbf{70.4} & \textbf{55.5} \\ \hline\hline
ResNet50 \cite{he2016deep}  & \multirow{4}{*}{\begin{tabular}[c]{@{}c@{}}Cascade\\ Mask\\ R-CNN\end{tabular} \cite{cai2018cascade}} & 46.3 & 64.3  & 50.5  & ResNet50 \cite{he2016deep}  & \multirow{4}{*}{ATSS \cite{zhang2020bridging}} & 43.5 & 61.9  & 47.0  \\
Swin-T \cite{liu2021swin}    &       & 50.5 & 69.3  & 54.9  & Swin-T  \cite{liu2021swin}   &           & 47.2 & 66.5  & 51.3  \\
PVTv2-B2 \cite{wang2021pvtv2}  &     & 51.1 & 69.8  & 55.3  & PVTv2-B2 \cite{wang2021pvtv2}  &         & 49.9 & 69.1  & 54.1  \\
Wave-ViT-S &                   & \textbf{52.1} & \textbf{70.7}  & \textbf{56.6}  & Wave-ViT-S &  & \textbf{50.7} & \textbf{69.8}  & \textbf{55.5}  \\ \Xhline{2\arrayrulewidth}
\end{tabular}
\label{table:od4}
\vspace{-0.2in}
\end{table*}

\subsubsection{Performance Comparison.}
Table \ref{table:od} lists the performance comparisons across different pre-trained vision backbones under the base detector of RetinaNet and Mask R-CNN for object detection and instance segmentation task, respectively. Note that we follow the evaluation for image recognition task by grouping all the pre-trained backbones into two categories (\emph{i.e.}, Small size and Base size). As shown in this table, the performance trends in each downstream task are similar to those in image recognition task. Concretely, under the similar model size for each group, the multi-scale ViT backbones (\emph{e.g.}, Swin-T/S and PVTv2-B2/B3) consistently exhibit better performances than CNN backbones (\emph{e.g.}, ResNet50/101) across all evaluation metrics. Furthermore, by capitalizing on wavelet transforms to enable lossless down-sampling in multi-scale self-attention learning, Wave-ViT variants outperform PVTv2-B2/B3 that explore sub-optimal down-sampling with pooling kernels. The results confirm that unifying self-attention learning and lossless down-sampling with wavelet transforms can improve the transfer capability of pre-trained multi-scale representations on dense prediction tasks.

To further verify the generalizability of the pre-trained multi-scale features via Wave-ViT for object detection, we evaluate various pre-trained vision backbones on four state-of-the-arts detectors (GFL, Sparse RCNN, Cascade Mask R-CNN, and ATSS). Table \ref{table:od4} shows the detailed performances of four object detectors with different pre-trained vision backbones under Small size. Similar to the observations in the base detector of RetinaNet, our Wave-ViT-S achieves consistent performance gains against both CNN backbone (ResNet50) and multi-scale ViT backbones (Swin-T and PVTv2-B2) across all the four state-of-the-arts detectors. This again validates the advantage of integrating multi-scale self-attention with invertible down-sampling in our Wave-ViT for object detection.

\begin{table*}[!tb]\scriptsize
\centering
\caption{The performances of various vision backbones on ADE20K validation dataset for the downstream task of semantic segmentation. We employ the commonly adopted base model (UPerNet) for semantic segmentation and report the mean IoU (mIoU) averaged over all classes for evaluation. We group all vision backbones into two categories: Small size and Base size.}
\vspace{-0.1in}
\begin{tabular}{cc|c|cc|c}
\Xhline{2\arrayrulewidth}
\multicolumn{3}{c|} {Small} & \multicolumn{3}{c} {Base} \\ \hline
Method       & Backbone                              & mIoU  & Method       & Backbone    & mIoU \\ \hline
UPerNet \cite{xiao2018unified}     & ResNet-50 \cite{he2016deep}           & 42.8  & UPerNet \cite{xiao2018unified}     & ResNet-101 \cite{he2016deep}             & 44.9 \\
UPerNet \cite{xiao2018unified}     & DeiT-S \cite{touvron2021training}     & 43.8  & UPerNet \cite{xiao2018unified}     & DeiT-B  \cite{touvron2021training}       & 47.2 \\
DeeplabV3 \cite{chen2018encoder}   & ResNeSt-50 \cite{zhang2020resnest}    & 45.1  & DeeplabV3 \cite{chen2018encoder}    & ResNeSt-101 \cite{zhang2020resnest}      & 46.9 \\
Semantic FPN \cite{kirillov2019panoptic} & PVTv2-B2 \cite{wang2021pvtv2}         & 45.2  & Semantic FPN \cite{kirillov2019panoptic} & PVTv2-B3 \cite{wang2021pvtv2}            & 47.3 \\
UPerNet \cite{xiao2018unified}     & RegionViT-S \cite{chen2021regionvit}  & 45.3  & UPerNet \cite{xiao2018unified}     & RegionViT-B \cite{chen2021regionvit}     & 47.5 \\
UPerNet \cite{xiao2018unified}     & Swin-T \cite{liu2021swin}             & 45.8  & UPerNet \cite{xiao2018unified}     & Twins-SVT-B \cite{chu2021twins}          & 48.9 \\
UPerNet \cite{xiao2018unified}     & Twins-SVT-S \cite{chu2021twins}       & 47.1  & UPerNet \cite{xiao2018unified}     & Swin-S \cite{liu2021swin}                & 49.5 \\
UPerNet \cite{xiao2018unified}     & Wave-ViT-S                            & \textbf{49.6}     & UPerNet \cite{xiao2018unified}     & Wave-ViT-B                               & \textbf{51.5}    \\ \Xhline{2\arrayrulewidth}
\end{tabular}
\label{table:ss}
\vspace{-0.3in}
\end{table*}

\subsection{Semantic Segmentation on ADE20K}
\subsubsection{Dataset and Optimization Setups.}
We next evaluate our pre-trained Wave-ViT in the downstream task of semantic segmentation on ADE20K dataset. This dataset is the most typical benchmark for evaluating semantic segmentation techniques, which consists of 25K images (20K training images, 2K validation images, and 3K testing images) derived from 150 semantic categories. Here we choose the commonly adopted UPerNet \cite{xiao2018unified} as the base model for this task and the CNN backbone in primary UPerNet is replaced with our Wave-ViT. During training, we train the models on 8 GPUs for 160K iterations via AdamW \cite{loshchilov2017decoupled} optimizer (batch size: 16, initial learning rate: 0.00006, weight decay: 0.01). Both linear learning rate decay scheduler and a linear warmup of 1,500 iterations are utilized for optimization. The scale of input images is fixed as 512 $\times$ 512. We perform the random horizontal flipping, random photometric distortion, and random re-scaling within the ratio range [0.5, 2.0] as data augmentations. We report the metric of mean IoU (mIoU) averaged over all classes for evaluation. For fair comparison with other vision backbones for semantic segmentation downstream task, we set all the hyperparameters and detection heads as in Swin \cite{liu2021swin}.

\subsubsection{Performance Comparison.}

Table \ref{table:ss} shows the mIoU scores of different pre-trained vision backbones under the base models (\emph{e.g.}, UPerNet, DeeplabV3, Semantic FPN) for semantic segmentation. As in the evaluation for image recognition, object detection, and instance segmentation tasks, we group all the pre-trained backbones into two categories (\emph{i.e.}, Small size and Base size). Similarly, by upgrading multi-scale self-attention learning with Wavelet based invertible down-sampling, our Wave-ViT variants yield consistent gains against both CNN backbones (\emph{e.g.}, ResNet-50/101 and ResNeSt-50/101) and existing multi-scale ViT backbones (\emph{e.g.}, Swin-T/S and Twins-SVT-S/B). Concretely, under a comparable model size within each group, Wave-ViT-S/B achieves 49.6\%/51.5\% mIoU on ADE20K validation dataset, which absolutely improves the best competitor Twins-SVT-S/Swin-S (47.1\%/49.5\%) with 2.5\%/2.0\%. The results basically demonstrates the superiority of Wave-ViT for semantic segmentation task.

\begin{figure}[!tb]
\vspace{-0.1in}
\centering {\includegraphics[width=0.52\textwidth]{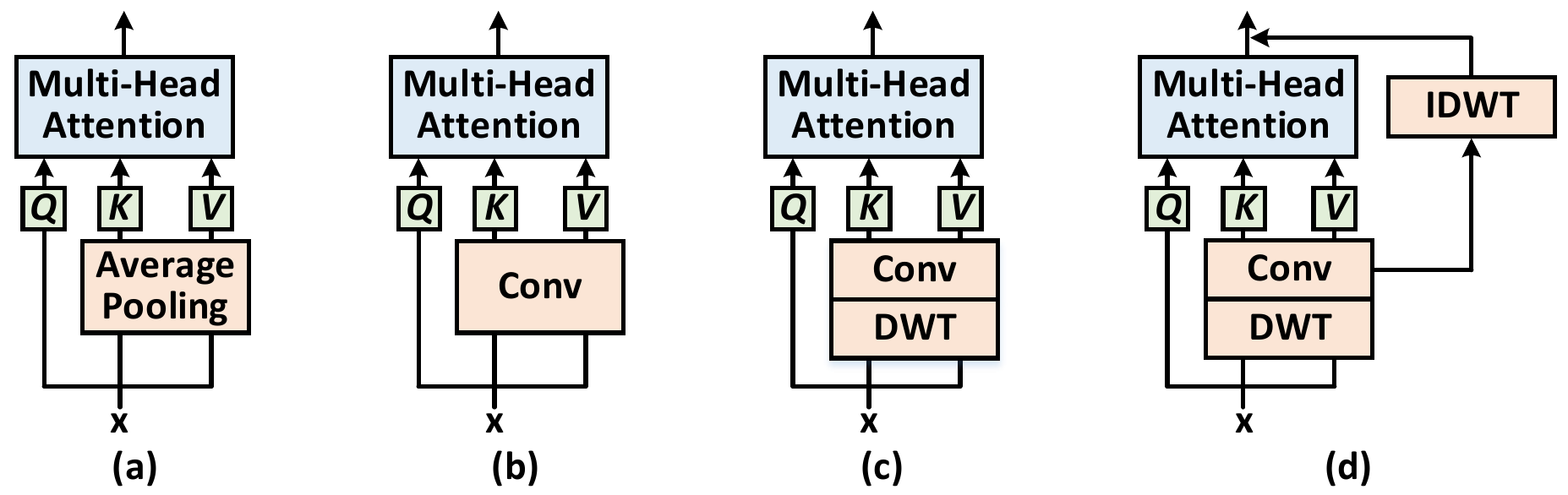}}
\qquad
\begin{tabular}[b]{c|cc|cc}
\hline
         &  Params  & GFLOPs  & Top-1 & Top-5 \\ \hline
  a      & 18.7M    & 3.9     & 82.0  & 96.0  \\
  b      & 20.4M    & 4.0     & 82.0  & 96.0  \\
  c      & 19.8M    & 4.3     & 82.5  & 96.1  \\
  d      & 19.8M    & 4.3     & \textbf{82.7}  & \textbf{96.2}  \\ \hline
\end{tabular}
\vspace{-0.16in}
\caption{Performance comparisons across different ways on designing self-attention blocks with down-sampling in multi-scale ViT backbones (under Small size): (a) self-attention block with irreversible down-sampling operation of average pooling, (b) self-attention block with irreversible down-sampling operation of pooling kernels, (c) a degraded version of Wavelets block that solely equips self-attention block with invertible down-sampling via wavelet transforms (DWT), and (d) the full version of our Wavelets block with inverse wavelet transforms (IDWT).}
\label{fig:ablation}
\vspace{-0.2in}
\end{figure}

\subsection{Ablation Study}

We investigate how each design in our Wavelets block influences the overall performance of multi-scale ViTs on ImageNet1K dataset for image recognition task, as summarized in Figure \ref{fig:ablation}. Note that all the variants of self-attention blocks here are constructed under similar model size (Small) for fair comparison.

\textbf{Block (a)} is a typical self-attention block with irreversible down-sampling. By directly operating average pooling over the input keys/values, (a) significantly reduces the computational cost for self-attention learning and the Top-1 score achieves 82.0\%.
\textbf{Block (b)} is another typical self-attention block with irreversible down-sampling via pooling kernels (convolution), rather than average pooling in (a). (b) reduces the spatial dimension of keys/values through pooling kernels, leading to the same performances as in (a). However, the number of parameters is inevitably increased.
\textbf{Block (c)} can be regarded as a degraded version of our Wavelets block, that solely equips self-attention block with invertible down-sampling based on wavelet transforms (DWT). Compared to the most efficient (a) with irreversible down-sampling, the Top-1 score of (c) increases from 82.0\% to 82.5\%. This validates the effectiveness of unifying self-attention block and invertible down-sampling without information dropping.
\textbf{Block (d)} (\emph{i.e.}, the full version of Wavelets block) further upgrades (c) by additionally exploiting inverse wavelet transforms (IDWT) to strengthen outputs with enlarged receptive field. Such design leads to performance boosts in Top-1 and Top-5 scores, with negligible increase in computational cost/memory.

\begin{figure*}[!tb]
\vspace{-0.1in}
\centering {\includegraphics[width=0.9\textwidth]{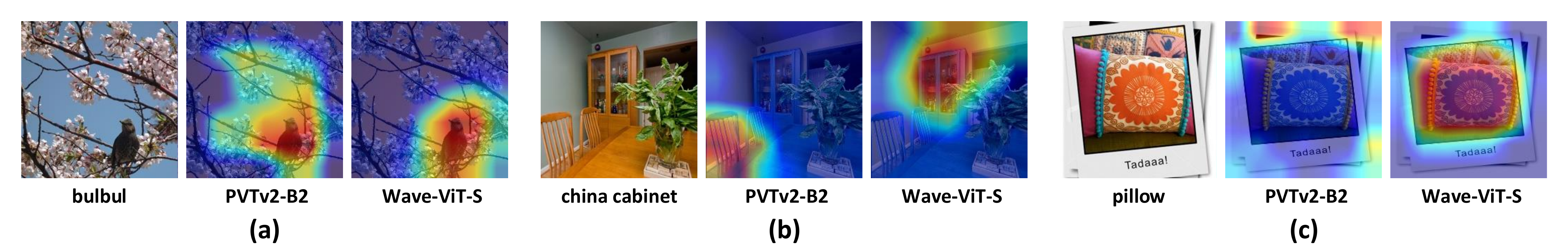}}
\vspace{-0.2in}
\caption{\small Visualization of Score-CAM \cite{wang2020score} for PVTv2-B2 \cite{wang2021pvtv2} and our Wave-ViT-S on six images in ImageNet1K dataset.}
\label{fig:figatt}
\vspace{-0.2in}
\end{figure*}

\subsection{Visualization of Learnt Visual Representation}
In order to further explain the visual representations learnt by our Wave-ViT, we produce the saliency map through Score-CAM \cite{wang2020score} to identify the importance of each pixel in presenting the class discrimination of the input image. Figure \ref{fig:figatt} visualizes the saliency map derived from the visual representations learnt by two backbones with similar model size (PVTv2-B2 and our Wave-ViT-S). As illustrated in the figure, Wave-ViT-S consistently shows higher concentration at the semantically relevant object than PVTv2-B2, which validates that the representations learnt by Wave-ViT-S are more robust.

\section{Conclusions}

In this paper, we delve into the idea of unifying typical Transformer module and invertible down-sampling, thereby pursuing efficient multi-scale self-attention learning with
lossless down-sampling. To verify our claim, we present a new principled Transformer module, \emph{i.e.}, Wavelets block, that capitalizes on Discrete Wavelet Transform (DWT) to perform invertible down-sampling over keys/values in self-attention learning. In addition, we adopt inverse DWT (IDWT) to reconstruct the down-sampled DWT outputs, which are utilized to strengthen the outputs of Wavelets block by aggregating local contexts with enlarged receptive field. Our Wavelets block is appealing in view that it is feasible to construct multi-scale ViT backbone with Wavelets blocks, with light computational cost/memory budget. In particular, by operating stacked Wavelets blocks over multi-scale features in four-stage architecture, a series of Wavelet Vision Transformer (Wave-ViT) are designed with different model sizes. We empirically validate the superiority of Wave-ViT over the state-of-the-art multi-scale ViT backbones for image recognition, under comparable numbers of parameters. Furthermore, our Wave-ViT also generalizes well to downstream tasks of object detection, instance segmentation, and semantic segmentation.
%
%
\bibliographystyle{splncs04}
\bibliography{egbib}

\begin{thebibliography}{10}
\providecommand{\url}[1]{\texttt{#1}}
\providecommand{\urlprefix}{URL }
\providecommand{\doi}[1]{https://doi.org/#1}

\bibitem{bae2017beyond}
Bae, W., Yoo, J., Chul~Ye, J.: Beyond deep residual learning for image
  restoration: Persistent homology-guided manifold simplification. In: CVPR
  Workshops (2017)

\bibitem{bello2019attention}
Bello, I., Zoph, B., Vaswani, A., Shlens, J., Le, Q.V.: Attention augmented
  convolutional networks. In: ICCV (2019)

\bibitem{cai2018cascade}
Cai, Z., Vasconcelos, N.: Cascade r-cnn: Delving into high quality object
  detection. In: CVPR (2018)

\bibitem{carion2020end}
Carion, N., Massa, F., Synnaeve, G., Usunier, N., Kirillov, A., Zagoruyko, S.:
  End-to-end object detection with transformers. In: ECCV (2020)

\bibitem{chen2021crossvit}
Chen, C.F., Fan, Q., Panda, R.: Crossvit: Cross-attention multi-scale vision
  transformer for image classification. In: ICCV (2021)

\bibitem{chen2021regionvit}
Chen, C.F., Panda, R., Fan, Q.: Regionvit: Regional-to-local attention for
  vision transformers. In: ICLR (2022)

\bibitem{chen2019mmdetection}
Chen, K., Wang, J., Pang, J., Cao, Y., Xiong, Y., Li, X., Sun, S., Feng, W.,
  Liu, Z., Xu, J., et~al.: Mmdetection: Open mmlab detection toolbox and
  benchmark. arXiv preprint arXiv:1906.07155  (2019)

\bibitem{chen2018encoder}
Chen, L.C., Zhu, Y., Papandreou, G., Schroff, F., Adam, H.: Encoder-decoder
  with atrous separable convolution for semantic image segmentation. In: ECCV
  (2018)

\bibitem{chu2021twins}
Chu, X., Tian, Z., Wang, Y., Zhang, B., Ren, H., Wei, X., Xia, H., Shen, C.:
  Twins: Revisiting the design of spatial attention in vision transformers. In:
  NeurIPS (2021)

\bibitem{cubuk2020randaugment}
Cubuk, E.D., Zoph, B., Shlens, J., Le, Q.V.: Randaugment: Practical automated
  data augmentation with a reduced search space. In: CVPR Workshops (2020)

\bibitem{d2021convit}
dAscoli, S., Touvron, H., Leavitt, M.L., Morcos, A.S., Biroli, G., Sagun, L.:
  Convit: Improving vision transformers with soft convolutional inductive
  biases. In: ICML (2021)

\bibitem{deng2009imagenet}
Deng, J., Dong, W., Socher, R., Li, L.J., Li, K., Fei-Fei, L.: Imagenet: A
  large-scale hierarchical image database. In: CVPR (2009)

\bibitem{dosovitskiy2020image}
Dosovitskiy, A., Beyer, L., Kolesnikov, A., Weissenborn, D., Zhai, X.,
  Unterthiner, T., Dehghani, M., Minderer, M., Heigold, G., Gelly, S., et~al.:
  An image is worth 16x16 words: Transformers for image recognition at scale.
  In: ICLR (2020)

\bibitem{fan2021multiscale}
Fan, H., Xiong, B., Mangalam, K., Li, Y., Yan, Z., Malik, J., Feichtenhofer,
  C.: Multiscale vision transformers. In: ICCV (2021)

\bibitem{fujieda2018wavelet}
Fujieda, S., Takayama, K., Hachisuka, T.: Wavelet convolutional neural
  networks. arXiv preprint arXiv:1805.08620  (2018)

\bibitem{gao2019res2net}
Gao, S.H., Cheng, M.M., Zhao, K., Zhang, X.Y., Yang, M.H., Torr, P.: Res2net: A
  new multi-scale backbone architecture. IEEE TPMAI  (2021)

\bibitem{glorot2010understanding}
Glorot, X., Bengio, Y.: Understanding the difficulty of training deep
  feedforward neural networks. In: AISTATS (2010)

\bibitem{guo2017deep}
Guo, T., Seyed~Mousavi, H., Huu~Vu, T., Monga, V.: Deep wavelet prediction for
  image super-resolution. In: CVPR Workshops (2017)

\bibitem{han2021transformer}
Han, K., Xiao, A., Wu, E., Guo, J., Xu, C., Wang, Y.: Transformer in
  transformer. In: NeurIPS (2021)

\bibitem{he2017mask}
He, K., Gkioxari, G., Doll{\'a}r, P., Girshick, R.: Mask r-cnn. In: ICCV (2017)

\bibitem{he2016deep}
He, K., Zhang, X., Ren, S., Sun, J.: Deep residual learning for image
  recognition. In: CVPR (2016)

\bibitem{hu2019local}
Hu, H., Zhang, Z., Xie, Z., Lin, S.: Local relation networks for image
  recognition. In: ICCV (2019)

\bibitem{hu2018squeeze}
Hu, J., Shen, L., Sun, G.: Squeeze-and-excitation networks. In: CVPR (2018)

\bibitem{jiang2021all}
Jiang, Z.H., Hou, Q., Yuan, L., Zhou, D., Shi, Y., Jin, X., Wang, A., Feng, J.:
  All tokens matter: Token labeling for training better vision transformers.
  In: NeurIPS (2021)

\bibitem{kirillov2019panoptic}
Kirillov, A., Girshick, R., He, K., Doll{\'a}r, P.: Panoptic feature pyramid
  networks. In: CVPR (2019)

\bibitem{krizhevsky2012imagenet}
Krizhevsky, A., Sutskever, I., Hinton, G.E.: Imagenet classification with deep
  convolutional neural networks. In: NeurIPS (2012)

\bibitem{lecun1998gradient}
LeCun, Y., Bottou, L., Bengio, Y., Haffner, P.: Gradient-based learning applied
  to document recognition. Proceedings of the IEEE  (1998)

\bibitem{li2020generalized}
Li, X., Wang, W., Wu, L., Chen, S., Hu, X., Li, J., Tang, J., Yang, J.:
  Generalized focal loss: Learning qualified and distributed bounding boxes for
  dense object detection. In: NeurIPS (2020)

\bibitem{li2022comprehending}
Li, Y., Pan, Y., Yao, T., Mei, T.: Comprehending and ordering semantics for
  image captioning. In: CVPR (2022)

\bibitem{li2022contextual}
Li, Y., Yao, T., Pan, Y., Mei, T.: Contextual transformer networks for visual
  recognition. IEEE TPAMI  (2022)

\bibitem{lin2017focal}
Lin, T.Y., Goyal, P., Girshick, R., He, K., Doll{\'a}r, P.: Focal loss for
  dense object detection. In: ICCV (2017)

\bibitem{lin2014microsoft}
Lin, T.Y., Maire, M., Belongie, S., Hays, J., Perona, P., Ramanan, D.,
  Doll{\'a}r, P., Zitnick, C.L.: Microsoft coco: Common objects in context. In:
  ECCV (2014)

\bibitem{liu2020wavelet}
Liu, L., Liu, J., Yuan, S., Slabaugh, G., Leonardis, A., Zhou, W., Tian, Q.:
  Wavelet-based dual-branch network for image demoir{\'e}ing. In: ECCV (2020)

\bibitem{liu2018multi}
Liu, P., Zhang, H., Zhang, K., Lin, L., Zuo, W.: Multi-level wavelet-cnn for
  image restoration. In: CVPR Workshops (2018)

\bibitem{liu2021swin}
Liu, Z., Lin, Y., Cao, Y., Hu, H., Wei, Y., Zhang, Z., Lin, S., Guo, B.: Swin
  transformer: Hierarchical vision transformer using shifted windows. In: ICCV
  (2021)

\bibitem{long2022stand}
Long, F., Qiu, Z., Pan, Y., Yao, T., Luo, J., Mei, T.: Stand-alone inter-frame
  attention in video models. In: CVPR (2022)

\bibitem{long2022DTF}
Long, F., Qiu, Z., Pan, Y., Yao, T., Ngo, C.W., Mei, T.: Dynamic temporal
  filtering in video models. In: ECCV (2022)

\bibitem{loshchilov2016sgdr}
Loshchilov, I., Hutter, F.: Sgdr: Stochastic gradient descent with warm
  restarts. arXiv preprint arXiv:1608.03983  (2016)

\bibitem{loshchilov2017decoupled}
Loshchilov, I., Hutter, F.: Decoupled weight decay regularization. arXiv
  preprint arXiv:1711.05101  (2017)

\bibitem{mallat1989theory}
Mallat, S.G.: A theory for multiresolution signal decomposition: the wavelet
  representation. IEEE TPAMI  (1989)

\bibitem{oyallon2017scaling}
Oyallon, E., Belilovsky, E., Zagoruyko, S.: Scaling the scattering transform:
  Deep hybrid networks. In: ICCV (2017)

\bibitem{pan2020x}
Pan, Y., Yao, T., Li, Y., Mei, T.: X-linear attention networks for image
  captioning. In: Proceedings of the IEEE/CVF Conference on Computer Vision and
  Pattern Recognition. pp. 10971--10980 (2020)

\bibitem{ramachandran2019stand}
Ramachandran, P., Parmar, N., Vaswani, A., Bello, I., Levskaya, A., Shlens, J.:
  Stand-alone self-attention in vision models. In: NeurIPS (2019)

\bibitem{simonyan2014very}
Simonyan, K., Zisserman, A.: Very deep convolutional networks for large-scale
  image recognition. In: ICLR (2015)

\bibitem{srinivas2021bottleneck}
Srinivas, A., Lin, T.Y., Parmar, N., Shlens, J., Abbeel, P., Vaswani, A.:
  Bottleneck transformers for visual recognition. In: CVPR (2021)

\bibitem{sun2021sparse}
Sun, P., Zhang, R., Jiang, Y., Kong, T., Xu, C., Zhan, W., Tomizuka, M., Li,
  L., Yuan, Z., Wang, C., et~al.: Sparse r-cnn: End-to-end object detection
  with learnable proposals. In: CVPR (2021)

\bibitem{szegedy2015going}
Szegedy, C., Liu, W., Jia, Y., Sermanet, P., Reed, S., Anguelov, D., Erhan, D.,
  Vanhoucke, V., Rabinovich, A.: Going deeper with convolutions. In: CVPR
  (2015)

\bibitem{touvron2021training}
Touvron, H., Cord, M., Douze, M., Massa, F., Sablayrolles, A., J{\'e}gou, H.:
  Training data-efficient image transformers \& distillation through attention.
  In: ICML (2021)

\bibitem{touvron2021going}
Touvron, H., Cord, M., Sablayrolles, A., Synnaeve, G., J{\'e}gou, H.: Going
  deeper with image transformers. In: ICCV (2021)

\bibitem{vaswani2017attention}
Vaswani, A., Shazeer, N., Parmar, N., Uszkoreit, J., Jones, L., Gomez, A.N.,
  Kaiser, {\L}., Polosukhin, I.: Attention is all you need. In: NeurIPS (2017)

\bibitem{wang2020score}
Wang, H., Wang, Z., Du, M., Yang, F., Zhang, Z., Ding, S., Mardziel, P., Hu,
  X.: Score-cam: Score-weighted visual explanations for convolutional neural
  networks. In: CVPR workshops (2020)

\bibitem{wang2020deep}
Wang, J., Sun, K., Cheng, T., Jiang, B., Deng, C., Zhao, Y., Liu, D., Mu, Y.,
  Tan, M., Wang, X., et~al.: Deep high-resolution representation learning for
  visual recognition. IEEE TPAMI  (2020)

\bibitem{wang2021scaled}
Wang, P., Wang, X., Luo, H., Zhou, J., Zhou, Z., Wang, F., Li, H., Jin, R.:
  Scaled relu matters for training vision transformers. arXiv preprint
  arXiv:2109.03810  (2021)

\bibitem{wang2021pvtv2}
Wang, W., Xie, E., Li, X., Fan, D.P., Song, K., Liang, D., Lu, T., Luo, P.,
  Shao, L.: Pvtv2: Improved baselines with pyramid vision transformer. arXiv
  preprint arXiv:2106.13797  (2021)

\bibitem{wang2021pyramid}
Wang, W., Xie, E., Li, X., Fan, D.P., Song, K., Liang, D., Lu, T., Luo, P.,
  Shao, L.: Pyramid vision transformer: A versatile backbone for dense
  prediction without convolutions. In: ICCV (2021)

\bibitem{williams2018wavelet}
Williams, T., Li, R.: Wavelet pooling for convolutional neural networks. In:
  ICLR (2018)

\bibitem{wu2021cvt}
Wu, H., Xiao, B., Codella, N., Liu, M., Dai, X., Yuan, L., Zhang, L.: Cvt:
  Introducing convolutions to vision transformers. In: ICCV (2021)

\bibitem{xiao2018unified}
Xiao, T., Liu, Y., Zhou, B., Jiang, Y., Sun, J.: Unified perceptual parsing for
  scene understanding. In: ECCV (2018)

\bibitem{xie2017aggregated}
Xie, S., Girshick, R., Doll{\'a}r, P., Tu, Z., He, K.: Aggregated residual
  transformations for deep neural networks. In: CVPR (2017)

\bibitem{xu2021co}
Xu, W., Xu, Y., Chang, T., Tu, Z.: Co-scale conv-attentional image
  transformers. In: ICCV (2021)

\bibitem{yang2021focal}
Yang, J., Li, C., Zhang, P., Dai, X., Xiao, B., Yuan, L., Gao, J.: Focal
  self-attention for local-global interactions in vision transformers. arXiv
  preprint arXiv:2107.00641  (2021)

\bibitem{yuan2021tokens}
Yuan, L., Chen, Y., Wang, T., Yu, W., Shi, Y., Jiang, Z.H., Tay, F.E., Feng,
  J., Yan, S.: Tokens-to-token vit: Training vision transformers from scratch
  on imagenet. In: ICCV (2021)

\bibitem{yuan2021volo}
Yuan, L., Hou, Q., Jiang, Z., Feng, J., Yan, S.: Volo: Vision outlooker for
  visual recognition. arXiv preprint arXiv:2106.13112  (2021)

\bibitem{zhang2020resnest}
Zhang, H., Wu, C., Zhang, Z., Zhu, Y., Zhang, Z., Lin, H., Sun, Y., He, T.,
  Mueller, J., Manmatha, R., et~al.: Resnest: Split-attention networks. arXiv
  preprint arXiv:2004.08955  (2020)

\bibitem{zhang2021token}
Zhang, H., Hao, Y., Ngo, C.W.: Token shift transformer for video
  classification. In: ACM Multimedia (2021)

\bibitem{zhang2019making}
Zhang, R.: Making convolutional networks shift-invariant again. In: ICML (2019)

\bibitem{zhang2020bridging}
Zhang, S., Chi, C., Yao, Y., Lei, Z., Li, S.Z.: Bridging the gap between
  anchor-based and anchor-free detection via adaptive training sample
  selection. In: CVPR (2020)

\bibitem{zhang2022exploring}
Zhang, Y., Pan, Y., Yao, T., Huang, R., Mei, T., Chen, C.W.: Exploring
  structure-aware transformer over interaction proposals for human-object
  interaction detection. In: CVPR (2022)

\bibitem{zhao2020exploring}
Zhao, H., Jia, J., Koltun, V.: Exploring self-attention for image recognition.
  In: CVPR (2020)

\bibitem{zhong2020random}
Zhong, Z., Zheng, L., Kang, G., Li, S., Yang, Y.: Random erasing data
  augmentation. In: AAAI (2020)

\bibitem{zhou2019semantic}
Zhou, B., Zhao, H., Puig, X., Xiao, T., Fidler, S., Barriuso, A., Torralba, A.:
  Semantic understanding of scenes through the ade20k dataset. IJCV  (2019)

\bibitem{zhou2021deepvit}
Zhou, D., Kang, B., Jin, X., Yang, L., Lian, X., Jiang, Z., Hou, Q., Feng, J.:
  Deepvit: Towards deeper vision transformer. arXiv preprint arXiv:2103.11886
  (2021)

\end{thebibliography}
\end{document}